\documentclass[conference,10pt]{IEEEtran}
\IEEEoverridecommandlockouts                          
\usepackage[font=small]{caption}
\usepackage[utf8]{inputenc}
\usepackage[T1]{fontenc}
\usepackage{graphicx}
\usepackage{subcaption} 
\usepackage{float}
\usepackage{placeins}
\usepackage{amssymb}  
\usepackage{amsmath}
\usepackage{flushend}
\usepackage{booktabs}
\usepackage{cite}
\usepackage{hyperref}
\hypersetup{hidelinks}
\usepackage{tikz}
\usepackage{pgfplots}
\usepackage{graphicx} 
\pgfplotsset{compat=1.18}
\usepackage{xcolor}
\colorlet{black}{black}
\usepackage{xcolor}

\definecolor{myorange}{RGB}{255,127,14}
\definecolor{myblue}{RGB}{31,119,180}
\definecolor{mygreen}{RGB}{44,160,44}
\definecolor{myred}{RGB}{214,39,40}

\title{\LARGE \bf
CBANet: A Compact Attention-Based CNN–BiLSTM Network for Aggressive Driving Event Detection}

\author{
Hanadi Alhamdan\textsuperscript{1,2},
Ghadah Alosaimi\textsuperscript{3,2}
Amir Atapour-Abarghouei\textsuperscript{2},
Farshad Arvin\textsuperscript{2} \\[0.5em]
\textsuperscript{1}Department of Computer Science, Princess Nourah bint Abdulrahman University, Saudi Arabia \\
\textsuperscript{2}Department of Computer Science, Durham University, UK \\
\textsuperscript{3}Department of Computer Science, Imam Mohammad Ibn Saud Islamic University, Saudi Arabia \\
}

\begin{document}

\maketitle

\begin{abstract}

Aggressive driving is a major cause of traffic accidents and poses a serious threat to road safety. Although deep learning methods have shown promising results in detecting risky driving behaviours from vehicle sensor data, their performance in real-world conditions is often limited by severe data imbalance, large variability between drivers, and the lack of physically interpretable vehicle dynamics representations. In this paper, we propose an enhanced deep learning framework for aggressive driving detection using multivariate vehicle dynamics signals. Instead of relying solely on raw measurements, the proposed approach constructs engineered dynamic features that capture steering, acceleration, and braking behaviour. To address the extreme rarity of aggressive events in naturalistic driving data, we introduce a stable training strategy that combines controlled SMOTE-based oversampling with a class-weighted loss formulation, and evaluates focal loss variants for imbalance handling. Furthermore, a safety-oriented decision strategy based on class-specific threshold calibration is adopted to better reflect the asymmetric risks of missed detections and false alarms in real-world applications. The proposed framework is evaluated on a newly collected naturalistic driving dataset. Extensive experiments show that the proposed method consistently outperforms standard deep learning baselines with significant improvements in minority-class recall and safety-critical F-score metrics while maintaining practical computational efficiency. Code: \url {https://github.com/halhamdan/CBANet}

\end{abstract}

\begin{IEEEkeywords}
Dynamic Neural Networks, Efficient and Tiny Neural Networks, Industrial applications, Neural Network Applications.
\end{IEEEkeywords}

\section{Introduction}
\label{sec:introduction}

Aggressive driving has long been recognized as a leading contributor to road accidents and close-call situations \cite{nhtsa2020}. The ability to detect such behaviour through vehicle-mounted sensors has become increasingly important for modern driver-assistance technologies, usage-based insurance programs, and intelligent transportation infrastructure \cite{ castignani2015driver}. Thanks to advances in automotive instrumentation, today's vehicles generate high-resolution driving data \cite{engstrom2005effects}, making it possible to apply learning-based solutions to recognize hazardous driving patterns automatically \cite{wang2022review}. Despite these technological advances, deploying such systems under real-world driving conditions presents considerable challenges \cite{johnson2011driving}.

The nature of real-world driving data introduces several complications. Aggressive events occur infrequently \cite{paefgen2013evaluation}, exhibit substantial variation from one instance to another \cite{martinez2018driving}, and frequently resemble normal driving behaviour \cite{eren2012estimating}. These characteristics result in heavily skewed datasets where distinguishing safe from unsafe driving becomes problematic \cite{he2009learning}. Existing approaches encounter three fundamental limitations. First, many models learn directly from raw sensor measurements without incorporating knowledge of vehicle dynamics, which can produce systems that perform inconsistently across different drivers, vehicles, and traffic scenarios \cite{kaur2019systematic}. Second, techniques that address data imbalance solely through resampling or modified loss functions often fall short when the imbalance becomes extreme \cite{huang2016learning}. Third, relying on conventional accuracy measures proves inadequate for safety-critical applications, where the trade-off between missing dangerous events and triggering false alarms must be carefully managed \cite{ davis2006relationship}.

To address these limitations, this paper proposes CBANet, a deep learning framework for aggressive driving recognition from multivariate vehicle dynamics time series. The framework integrates physically interpretable vehicle dynamics features, imbalance-aware training, and class-specific decision calibration.

The main contributions of this paper are summarised as follows:
\begin{itemize}

\item We propose a unified pipeline for aggressive driving recognition that combines dynamic feature construction, temporal modelling, and imbalance-aware learning.

\item We develop a training strategy for imbalanced naturalistic driving data using controlled synthetic minority oversampling technique (SMOTE) and class-weighted loss.

\item We introduce a class-specific decision and evaluation protocol for safety-oriented assessment of aggressive driving events.

\item We collect a new naturalistic driving dataset recorded in real-world traffic conditions, and extensive experiments on this dataset demonstrate the effectiveness and robustness of the proposed framework.
\end{itemize}
\section{Related Work}
\label{sec:related_work}

Considerable research has addressed aggressive driving detection by leveraging vehicle dynamics and computational learning techniques.  This section reviews relevant literature across two key themes: vehicle-dynamics-based event modelling, encompassing signals, features, and labelling practices (Section  \ref{subsec:AggEvent}), and the progression of learning methodologies from traditional classifiers to modern state-space models (Section \ref{subsec:LearnMeth}). These thematic areas provide context for positioning the contributions of the present work within the field of intelligent driver monitoring systems.

\subsection{Vehicle-Dynamics–Based Aggressive Event Modelling}
\label{subsec:AggEvent}
A consistent finding across the literature is that aggressive and unsafe events are well expressed in vehicle dynamics, especially longitudinal/lateral acceleration, speed, braking, and steering-related signals acquired from Controller Area Network (CAN) interfaces, On-Board Diagnostics (OBD) systems, Inertial Measurement Units (IMUs), or digital tachographs. Early studies utilized these movement indicators to detect speeding and sudden motion behaviours \cite{chenSpeedingBehaviorEstimation2018}, while later work expanded detection capabilities to include aggressive cornering, lane switching, rapid acceleration, and hard braking using multiple in-vehicle sensor streams \cite{alvarez-coelloModelingDangerousDriving2019}.
The dominant methodology in existing research uses an event-based framework, where aggressive driving is defined by particular actions including hard acceleration or braking, abrupt turns, and quick lane changes. These systems commonly employ physics-based threshold techniques to isolate and categorize driving events from continuous vehicle data \cite{alamriEffectiveBioSignalBasedDriver2020b,congExperimentalResearchDriver2021,  yukselDriversBlackBox2021, luBilevelDistributionMixture2021, liMacroscopicBigData2022, chandraUsingGraphTheoreticMachine2022, kimDesignImplementationOnDevice2023}. Although threshold-based approaches offer advantages in computational simplicity, transparency, and implementation ease, they face fundamental constraints regarding adaptability and portability, as their performance is heavily influenced by vehicle-specific features, sensor setups, and driving context.
Several works move beyond isolated event detection toward driver-level or macroscopic behaviour characterization, using metrics such as the frequency of harsh events per unit distance or time, as well as long-term aggregated statistics, particularly in the context of fleet management and insurance analytics \cite{luBilevelDistributionMixture2021, liMacroscopicBigData2022}. 
In parallel, a number of studies investigate multi-modal sensing strategies, where vehicle dynamics are complemented with inertial or wearable sensors in order to improve robustness and discriminative power \cite{alamriEffectiveBioSignalBasedDriver2020b, yukselDriversBlackBox2021}. 

\subsection{Learning Methodologies and Model Comparisons}
\label{subsec:LearnMeth}
Existing methodologies for aggressive driving detection can be categorized into three distinct paradigms. The first comprises classical machine learning pipelines that apply supervised classification algorithms, including Support Vector Machines (SVM), k-Nearest Neighbors (k-NN), decision trees, and Random Forests, to threshold-segmented or manually extracted events \cite{chenSpeedingBehaviorEstimation2018, marafieAutoCoachDrivingBehavior2019, congExperimentalResearchDriver2021, yukselDriversBlackBox2021}. These approaches offer computational efficiency and model interpretability but exhibit limited capacity for capturing complex temporal dependencies inherent in driving behaviour sequences.
The second paradigm leverages deep sequential architectures, particularly Convolutional Neural Networks (CNNs), Long Short-Term Memory networks (LSTMs), and Gated Recurrent Units (GRUs), which frame the detection problem as time series pattern recognition \cite{mumcuogluDriverEvaluationHeavy2020, leePrivacyPreservingLearningMethod2023b}. These models improve accuracy but can be computationally demanding or less scalable for long sequences.
The third emerging paradigm explores structured sequence models and graph-based formulations designed for interaction-aware and trajectory-level reasoning \cite{triratDFTARDeepFusion2021, haqueDrivingManeuverClassification2022, huDetectingSociallyAbnormal2023, triratMGTARMultiViewGraph2023}. While these approaches exhibit powerful representational capabilities, they often depend on external perception or reconstructed trajectories rather than purely on-board signals.

\section{Methodology}
\label{sec:methodology}

\subsection{Preliminaries}

In this work, aggressive driving event detection is formulated as a supervised multi-class classification problem over multivariate vehicle dynamics time series. A driving session is represented as a sequence of multivariate observations:
\begin{equation}
\mathbf{X} = \{ \mathbf{x}_t \in \mathbb{R}^{F} \mid t = 1, \ldots, T \},
\end{equation}
where $F$ denotes the number of signal channels, including both raw sensor measurements and engineered dynamic features, and $T$ is the sequence length. The objective is to learn a nonlinear mapping function:

\begin{equation}
f_\theta: \mathbb{R}^{W \times F} \rightarrow \mathcal{Y},
\end{equation}
where $W$ denotes the temporal window length and

\begin{equation}
\begin{aligned}
\mathcal{Y} = \{ & \text{Normal}, \text{Harsh Acceleration}, \\
                 & \text{Harsh Braking}, \text{Harsh Turning} \}
\end{aligned}
\end{equation}
is the set of driving event classes. Each output is therefore a probability distribution over multiple event categories, and the task is inherently a multi-class temporal pattern recognition problem rather than a binary or point-wise classification problem.

Instead of classifying individual time samples, the model operates on short temporal segments, each representing an event episode.

\subsection{Problem Description}

Given a multivariate time series segment:
\begin{equation}
\mathbf{X}_k = \{ \mathbf{x}_t \mid t = kS, \ldots, kS + W - 1 \},
\end{equation}
extracted using a sliding-window strategy with window length $W$ and stride $S$, the goal is to predict the driving behaviour class associated with this segment. Each segment is assigned a label based on the dominant event occurring within the window. This converts the original continuous driving signal into a set of overlapping event-centric temporal samples, each of which is treated as an independent training instance. The learning problem is therefore cast as a sequence-to-label mapping problem, where the model must infer the underlying driving intent from the temporal structure of multivariate vehicle dynamics signals.

\subsection{Model Architecture}

\begin{figure*}[t]
    \centering
    \includegraphics[width=\textwidth]{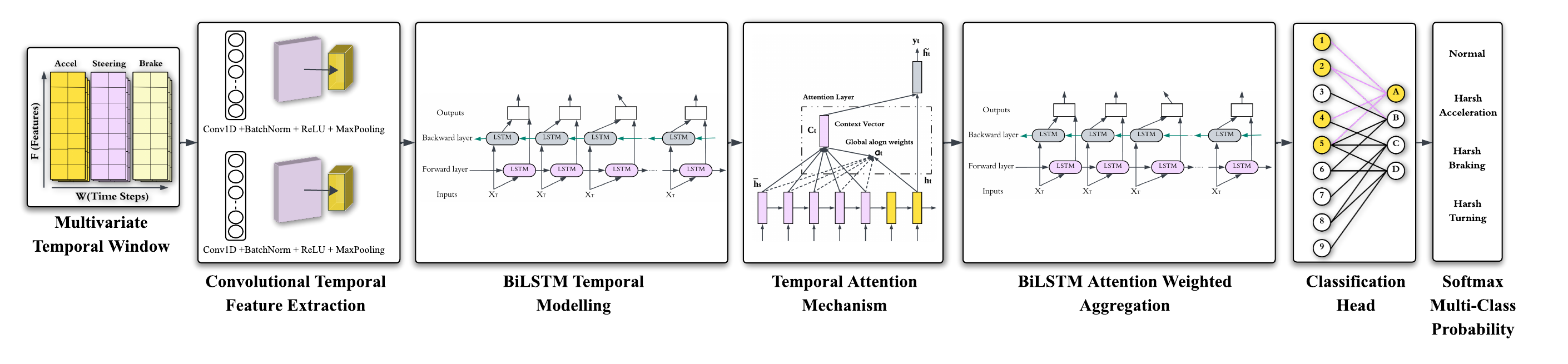}
    \caption{Proposed CBANet: a multivariate sliding window is encoded by a two-stage 1D CNN, modelled with BiLSTM and temporal attention, aggregated by a second BiLSTM, and classified via fully connected layers with Softmax into four driving events.}
    \label{fig:architecture}
\end{figure*}

The proposed network follows a deep hierarchical temporal modelling paradigm composed of five main components (Figure \ref{fig:architecture}): a two-stage one-dimensional convolutional (CNN) feature extractor, bidirectional long short-term memory (BiLSTM) temporal modelling layer with sequence output, an explicit temporal attention mechanism, a second BiLSTM for attention-weighted temporal aggregation, and a fully connected classification head with a softmax output. The network takes as input a multivariate temporal window $\mathbf{X} \in \mathbb{R}^{W \times F}$, where $W$ is the window length and $F$ is the number of features, and outputs a probability distribution over the driving behaviour classes. This design reflects the multi-scale nature of driving behaviours, which are characterized by short-term motion primitives, longer-term temporal dependencies, and a small number of highly discriminative critical moments.

\subsubsection{Convolutional Temporal Feature Extractor}

The first stage of the network consists of two stacked one-dimensional convolutional layers applied along the temporal dimension of the input window. Given an input segment $\mathbf{X} \in \mathbb{R}^{W \times F}$, the first convolutional transformation is defined as:
\begin{equation}
\mathbf{H}_1 = \sigma(\mathrm{BN}(\mathbf{X} * \mathbf{K}_1)),
\end{equation}
followed by a second convolutional layer:
\begin{equation}
\mathbf{H}_2 = \sigma(\mathrm{BN}(\mathbf{H}_1 * \mathbf{K}_2)),
\end{equation}
where $*$ denotes one-dimensional convolution, $\mathbf{K}_1 = 5$ and $\mathbf{K}_2 = 3$ are learnable kernel banks, $\mathrm{BN}(\cdot)$ denotes batch normalization, and $\sigma(\cdot)$ is the ReLU activation function. 1D max-pooling layers are applied between convolutional blocks with a pool size and stride of 2 to reduce temporal resolution and improve robustness to noise and small temporal misalignments.

In the implemented model, this CNN stage is explicitly multi-layer and multi-filter, allowing it to learn a hierarchy of temporal features. The early convolutional filters (64 filters) specialize in detecting very local motion primitives, such as sudden spikes in acceleration or brief oscillations in steering, while deeper filters (128 filters) combine these primitives into more complex short-term patterns. Batch normalization and dropout (0.2) are used to stabilize training and reduce overfitting, which is particularly important given the high correlation between neighbouring time samples in driving data.

\subsubsection{BiLSTM Temporal Modelling}

While convolutional layers are effective at capturing local temporal patterns, aggressive driving events are defined by their temporal evolution over several seconds. To explicitly model these longer-range dependencies, the resulting feature map of shape $(B, \tfrac{T}{4}, 128) $ is passed to a BiLSTM network. The forward and backward hidden states are computed as:
\begin{equation}
\overrightarrow{\mathbf{h}}_t = \mathrm{LSTM}_f(\mathbf{h}_t, \overrightarrow{\mathbf{h}}_{t-1}), \quad
\overleftarrow{\mathbf{h}}_t = \mathrm{LSTM}_b(\mathbf{h}_t, \overleftarrow{\mathbf{h}}_{t+1}),
\end{equation}
and the final representation at each time step is obtained by concatenation:
\begin{equation}
\mathbf{h}_t = [\overrightarrow{\mathbf{h}}_t ; \overleftarrow{\mathbf{h}}_t].
\end{equation}

The bidirectional structure is crucial in this application because the semantic meaning of an event often depends on both its onset and its recovery phase. Dropout is applied within the recurrent block to improve generalization and prevent overfitting to session-specific temporal patterns.

\subsubsection{Temporal Attention Mechanism}

Although the BiLSTM produces a sequence of hidden states describing the entire temporal window, not all time steps are equally informative for classification. To allow the network to automatically identify and emphasise the most discriminative moments, a temporal attention mechanism is applied. For each time step, an importance score is computed as:
\begin{equation}
e_t = \tanh(\mathbf{w}^\top \mathbf{h}_t + b),
\end{equation}
and normalized using a softmax function:
\begin{equation}
\alpha_t = \frac{\exp(e_t)}{\sum_{k=1}^{T} \exp(e_k)}.
\end{equation}
The final context vector is then obtained as a weighted sum of the hidden states:
\begin{equation}
\mathbf{c} = \sum_{t=1}^{T} \alpha_t \mathbf{h}_t.
\end{equation}

\subsubsection{Second BiLSTM Layer (Attention-Weighted Aggregation)} 

The attention-weighted sequence is passed to a second BiLSTM layer. This layer aggregates the attended temporal information into a compact, event-level representation 
that summarizes the entire window after attention focusing.

\subsubsection{Dense Layers and Softmax Multi-class Output}

The attention-aggregated context vector is passed through a stack of fully connected layers:
\begin{equation}
\mathbf{z}_1 = \sigma(\mathrm{BN}(\mathbf{W}_1 \mathbf{c} + \mathbf{b}_1)), \quad
\mathbf{z}_2 = \sigma(\mathbf{W}_2 \mathbf{z}_1 + \mathbf{b}_2).
\end{equation}
Finally, a softmax layer produces the multi-class probability distribution:
\begin{equation}
\hat{\mathbf{y}} = \mathrm{softmax}(\mathbf{W}_o \mathbf{z}_2 + \mathbf{b}_o).
\end{equation}
The network is trained using an imbalance-aware loss function, which is described in detail in Section ~\ref{subsec:focal_loss}.

\subsection{Loss Function (Class-Weighted and Focal Variants)} 
\label{subsec:focal_loss}

To address the pronounced class imbalance in aggressive driving events, the network is trained using a class-weighted loss formulation. In addition, a focal loss variant is considered an extension to further emphasise hard and misclassified samples. Let $C$ denote the number of classes, $y_i \in \{0, \ldots, C-1\}$ the ground-truth label for sample $i$, and $\hat{\mathbf{p}}_i = (\hat{p}_{i,1}, \ldots, \hat{p}_{i,C})$ the Softmax output vector. We define:
\begin{equation}
p_{t,i} = \hat{p}_{i, y_i}
\end{equation}
as the predicted probability of the true class.

The loss for a single sample is defined as:
\begin{equation}
\mathcal{L}_i = - \alpha_{y_i} \, (1 - p_{t,i})^{\gamma_{y_i}} \, \log(p_{t,i}),
\end{equation}
and the batch loss is computed as the mean over all samples:
\begin{equation}
\mathcal{L} = \frac{1}{N} \sum_{i=1}^{N} \mathcal{L}_i .
\end{equation}

Here, $\alpha_{y_i}$ is a class weight that compensates for frequency imbalance, and $\gamma_{y_i}$ is a class-specific focusing parameter that down-weights easy examples and emphasises hard, misclassified samples. In the implementation, $\alpha$ is derived from the post-augmentation class weights and normalised, while $\gamma$ is treated as a tunable parameter and varied in the ablation study to analyse its effect on model performance.

Compared to standard class-weighted cross-entropy, this formulation introduces a modulating factor that down-weights well-classified samples and emphasises harder examples. However, as demonstrated in the ablation study (Section ~\ref{subsec:ablation}), strong focal modulation is not required in the proposed framework, as class imbalance is already effectively mitigated through controlled SMOTE oversampling and class weighting.

\section{Experimental Design}
\label{sec:ExpDesign}

\subsection{Dataset}

\begin{figure}[t]
\centering
\includegraphics[width=\linewidth]{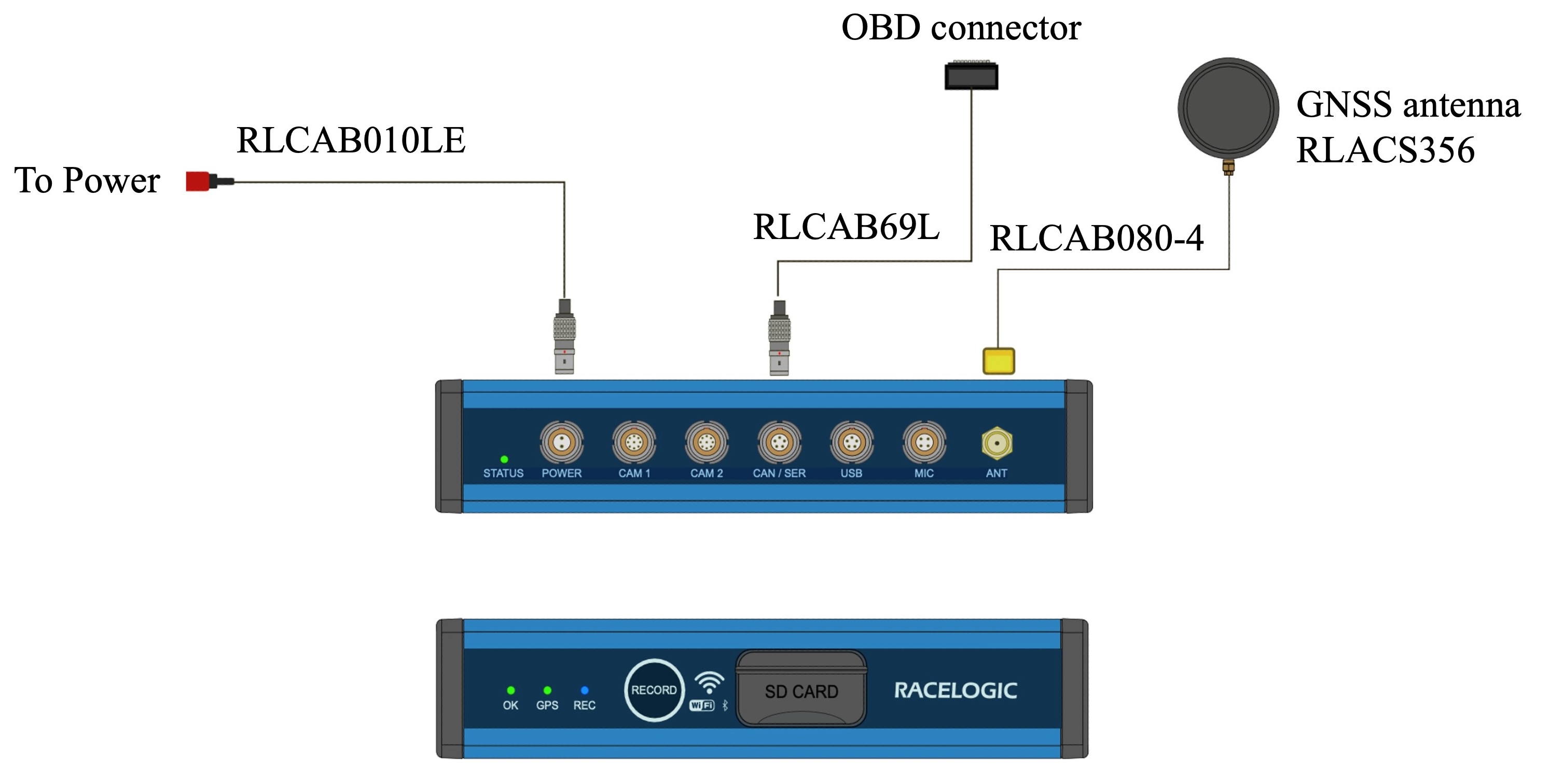}
\caption{Racelogic VBOX Video HD2 setup showing power, OBD-II, CAN/serial, and GNSS antenna connections for data collection.}
\label{fig:VBOX}
\end{figure}

The aggressive driving behaviour dataset is collected in real-world traffic contexts from 20 drivers (12 males, 8 females) aged between approximately 20 and 55 years, and each participant completed four driving sessions with an average duration of 40 minutes per session. Data were collected using a Renault Koleos (2018) instrumented vehicle. Vehicle dynamics data were recorded using a Racelogic VBOX Video HD2 data logger interfaced with the vehicle’s CAN bus via the OBD-II port, capturing high-resolution telemetry signals including speed, longitudinal and lateral acceleration, and selected power train parameters. In addition, global positioning data were acquired through an external GNSS antenna to enable precise localization and spatiotemporal alignment of driving events. All signals were synchronously sampled at 25 Hz. The VBOX is shown in Fig. ~\ref{fig:VBOX}.

Data collection was conducted along a predefined urban route of approximately 17 km, to ensure consistency in road geometry, traffic control, and environmental conditions across all sessions. The route comprised a mixture of urban segments, signalized intersections, and multi-lane arterial roads, representing typical everyday driving. Two sessions were conducted during Friday morning off-peak periods (07:00–10:00) to capture baseline driving under low traffic, while two additional sessions were carried out during weekday midday peak hours (14:00–17:00) to reflect dense traffic conditions. This design exposed drivers to both low-traffic and dense-traffic conditions, including frequent acceleration–deceleration cycles, lane changes, and interactions with dense and occasionally unpredictable traffic.
\subsection{Data Labelling}
\label{subsec:data_labelling}

Reliable supervision is essential for aggressive driving detection. In this work, ground-truth labels are generated using a deterministic event detection pipeline based on vehicle dynamics that operates directly on synchronized vehicle telemetry signals \cite{lindow2020ai}. Each time sample is assigned to one of four mutually exclusive classes: Normal, Harsh Acceleration, Harsh Braking, and Harsh Turning.
The procedure is designed to be conservative, reproducible, and physically interpretable, avoiding driver-relative or distribution-dependent definitions of aggressiveness.

The threshold values are selected based on established ranges reported in prior literature on aggressive driving behaviour \cite{lindow2020ai}, which associates longitudinal accelerations beyond $\pm0.3$–$0.4g$ and lateral accelerations above $0.5g$ with harsh events. In this work, these thresholds are slightly adjusted to $\theta_{\text{brake}} = -0.35g$, $\theta_{\text{accel}} = 0.38g$, and $\theta_{\text{turn}} = 0.55g$ to better align with the noise characteristics and sampling resolution of the collected dataset.

To suppress noise and capture events as short temporal events rather than isolated spikes, all signals are processed using a sliding temporal window of duration $W_s = 4$~s, and rolling extrema are computed as:
\begin{equation}
a_{\text{long}}^{\max}(t) = \max_{\tau \in [t-W_s,\, t]} a_{\text{long}}(\tau), \quad
a_{\text{lat}}^{\max}(t) = \max_{\tau \in [t-W_s,\, t]} |a_{\text{lat}}(\tau)|.
\end{equation}

Harsh braking is detected using a dual physical validation that combines both longitudinal acceleration and speed-derived deceleration. Let:
\begin{equation}
a_{\text{speed}}(t) = \frac{v(t) - v(t-1)}{\Delta t},
\end{equation}
and let $a_{\text{speed}}^{\min}(t)$ denote its rolling minimum over the same temporal window. A braking event is declared when $v(t) > v_{\min}$, $a_{\text{speed}}^{\min}(t) \le \theta_{\text{speed}}$, and $a_{\text{long}}(t) \le \theta_{\text{brake}}$, where $v_{\min} = 15~\text{km/h}$, with an additional intent-based rescue condition using the brake pedal signal to capture meaningful but slightly sub-threshold events. 
Harsh acceleration is detected when $v(t) > v_{\min}$, $a_{\text{long}}^{\max}(t) \ge \theta_{\text{accel}}$, throttle intent is present, and braking is not active. 
Harsh turning is detected independently using a higher speed gate, $v(t) > 30~\text{km/h}$, and the condition $a_{\text{lat}}^{\max}(t) \ge \theta_{\text{turn}}$.

To enforce temporal coherence, all detected events are post-processed using morphological closing, minimum-duration filtering ($T_{\min} = 0.4$~s), and temporal expansion to capture ramp-up and recovery phases, followed by a second duration filtering stage. Since multiple conditions may be satisfied simultaneously, conflicts are resolved using a fixed priority order: turning $>$ braking $>$ acceleration $>$ normal. The final label assignment is therefore given by:
\begin{equation}
y(t) =
\begin{cases}
\text{Harsh Turning}, & \text{if } \mathcal{T}(t), \\
\text{Harsh Braking}, & \text{else if } \mathcal{B}(t), \\
\text{Harsh Acceleration}, & \text{else if } \mathcal{A}(t), \\
\text{Normal}, & \text{otherwise}.
\end{cases}
\end{equation}

To validate the reliability of the proposed labelling pipeline, a controlled verification procedure was conducted. A subset of driving sessions was collected in which drivers were instructed to intentionally perform predefined events, including harsh braking, harsh acceleration, and harsh turning, under safe conditions. During data collection, the approximate timestamps of these intentional events were manually recorded to provide reference annotations. The proposed labelling pipeline was then applied to the same data without modification.

The automatically generated labels were compared against the manually recorded event timestamps. The results showed a high degree of temporal alignment between detected and intentional events, confirming that the threshold-based labelling strategy accurately captures meaningful aggressive behaviours.

\subsection{Data Preprocessing Pipeline}
\label{subsec:preprocessing}

Prior to model training, all vehicle dynamics signals are first cleaned and transformed into a set of physically interpretable and event-specific features.

\subsubsection{Feature Engineering}

Let $b(t)$ denote brake pedal position, the deceleration component of longitudinal acceleration is defined as:
\begin{equation}
a_{\text{long}}^{-}(t) = \max\big(0, -a_{\text{long}}(t)\big).
\end{equation}

Steering behaviour is characterized using two features derived from lateral acceleration. The instantaneous steering sharpness is measured by:
\begin{equation}
\text{TurnSharpness}(t) = \left| a_{\text{lat}}(t) - a_{\text{lat}}(t-1) \right|,
\end{equation}
which captures abrupt steering changes. Sustained turning is represented by a smoothed lateral acceleration signal:
\begin{equation}
a_{\text{lat}}^{\text{smooth}}(t) = \frac{1}{5} \sum_{k=0}^{4} a_{\text{lat}}(t-k).
\end{equation}

Braking aggressiveness is modelled using two complementary features. The physical severity of braking is quantified by:
\begin{equation}
P_{\text{decel}}(t) = v(t) \cdot a_{\text{long}}^{-}(t),
\end{equation}
which accounts for speed-dependent energy dissipation. Driver braking intent is captured by:
\begin{equation}
B_{\text{engage}}(t) = b(t) \cdot a_{\text{long}}^{-}(t),
\end{equation}
which fuses brake input with measured deceleration to distinguish active braking from passive deceleration.

\subsubsection{Signal Cleaning and Normalization}

Each feature channel is standardized using z-score normalization:
\begin{equation}
\tilde{x}^{(f)}(t) = \frac{x^{(f)}(t) - \mu_f}{\sigma_f},
\end{equation}
where $\mu_f$ and $\sigma_f$ are the mean and standard deviation of feature $f$ computed on the training set only, in order to avoid information leakage.

\subsubsection{Sliding-Window Segmentation}

Finally, the augmented multivariate signal is segmented into overlapping fixed-length windows using a sliding-window strategy, converting the continuous stream into event-centric temporal segments that serve as training samples. This is necessary because aggressive driving behaviours manifest over short temporal intervals rather than at isolated instants, enabling the model to learn from their temporal structure.

The class label of each window is determined using a label-safe voting scheme that prioritises aggressive events. Specifically, if any sustained aggressive behaviour is detected within the window after temporal consistency enforcement, the window is assigned to that class. Otherwise, majority voting is applied. To further suppress label noise and maintain consistency with underlying vehicle dynamics, a rule-based override mechanism is introduced: windows exhibiting extremely strong braking, turning, or acceleration patterns according to interpretable thresholds on deceleration, lateral acceleration, pedal engagement, and speed are reassigned to the corresponding aggressive class. This hybrid strategy ensures that short but critical events are not diluted by surrounding normal driving samples.

\subsection{Model Training and Evaluation}

The resulting windowed dataset is split into training, validation, and test subsets using stratified sampling to preserve the class distribution across all partitions. As is typical in naturalistic driving data, the class distribution is highly imbalanced, with normal driving dominating. To mitigate this, the training set is augmented using a controlled oversampling strategy based on SMOTE \cite{chawla2002smote}. Each window is temporarily flattened into a vector representation, and minority classes are oversampled up to a fixed fraction of the majority class size, while the majority class itself is never downsampled. This procedure significantly improves class balance while preserving the temporal structure of the data.

 Although SMOTE substantially reduces imbalance, a mild degree of class skew may remain; therefore, class weights are computed from the training distribution and slightly increased for safety-critical classes such as harsh braking and harsh turning. These weights are incorporated into a class-weighted loss formulation. While a focal loss variant is also evaluated to further emphasise hard samples, experimental results indicate that a simpler weighted cross-entropy formulation provides more stable and effective performance in this setting.

The network is optimized using the AdamW optimizer with decoupled weight decay regularization. Training is performed for up to 300 epochs using mini-batches with early stopping based on validation loss to prevent overfitting and automatic learning rate reduction when validation performance plateaus. The best-performing model parameters according to the validation loss are retained.

All experiments are evaluated on a strictly held-out test set. Since failing to detect aggressive events is more critical than producing false alarms in safety-oriented driving applications, overall performance is primarily summarised using the $F_2$-score, which assigns higher weight to recall than to precision. Both macro-averaged and class-frequency-weighted $F_2$-scores are reported in order to provide complementary perspectives on balanced performance and performance under the natural class distribution.
\section{Experimental Results}
\label{sec:Results}

The following experiments are evaluated: a comparison with classical and deep learning baselines, a computational efficiency analysis, an evaluation at session-level and driver-level, as well as an ablation study. To exclude the influence of the model training environment, all experiments are conducted on the same device with an NVIDIA RTX A6000 GPU. The window size is set to $W = 4$ seconds, and the prediction horizon is set to $H = 0 $ seconds.

\subsection{Comparison with Classical and Deep Learning Baselines}


\begin{table}[!t]
\centering
\caption{Comparison with Classical, Deep Sequential, and Graph-Based Models}
\label{tab:baseline_comparison}
\renewcommand{\arraystretch}{0.92} 
\begin{tabular}{lccc}
\toprule
\textbf{Model} & \textbf{Accuracy} & \textbf{ROC-AUC} & \textbf{F2 (Weighted)} \\
\midrule
\multicolumn{4}{l}{\textbf{Classical ML}} \\
SVM   & 0.9225 & 0.9854 & 0.9230 \\
K-NN  & 0.8806 & 0.9682 & 0.8824 \\
\midrule
\multicolumn{4}{l}{\textbf{Deep Sequential}} \\
CNN   & 0.9386 & 0.9899 & 0.9385 \\
LSTM  & 0.9413 & 0.9893 & 0.9413 \\
GRU   & 0.9416 & 0.9907 & 0.9414 \\
\midrule
\multicolumn{4}{l}{\textbf{Graph / Interaction}} \\
RGAT (GAT + GRU) & 0.8947 & 0.9781 & 0.8947 \\
GCN   & 0.7005 & 0.8767 & 0.8953 \\
GAT   & 0.8945 & 0.9774 & 0.8953 \\
\midrule
\textbf{CBANet} & \textbf{0.9585} & \textbf{0.9928} & \textbf{0.9584} \\
\bottomrule
\end{tabular}
\vspace{-1.5mm}
\end{table}

\begin{figure}[!t]
    \centering
    \begin{minipage}[b]{\linewidth}
        \centering
        \includegraphics{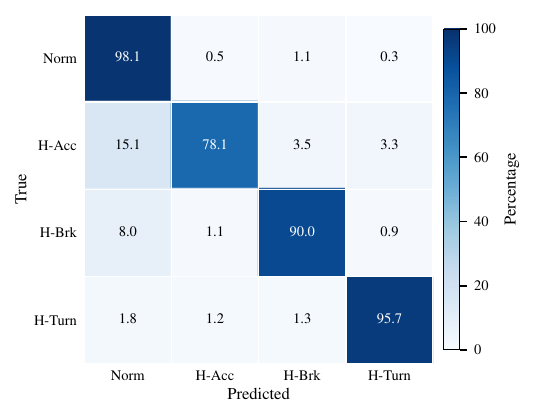}
    \end{minipage}
   
    \caption{Confusion matrix of CBANet.}
    \label{fig:cbanet_metrics}
\end{figure}

To evaluate the effectiveness of our proposed model, we conducted a comprehensive comparison against a wide range of baselines across different model categories. These include classical machine learning models like SVM, and KNN, deep sequential models such as CNN, LSTM, and GRU, and graph-based models like GCN, GAT, and a GRU-enhanced GAT variant. These baselines cover a diverse range of modelling strategies commonly applied to time series and behavioural data. As shown in Table ~\ref{tab:baseline_comparison}, our model (CBANet) consistently outperforms these approaches in detecting driving behaviours, demonstrating its ability to capture temporal dynamics more effectively. CBANet’s confusion matrix is illustrated in Fig.~\ref{fig:cbanet_metrics}.

\subsection{Computational Efficiency Analysis}

To support the compactness of the proposed architecture, we evaluate its parameter count, model size, and inference latency. CBANet contains only 180,325 parameters, with a storage footprint of approximately 0.76 MB, enabling efficient deployment on resource-constrained platforms.

The model achieves an average inference time of 1.43 ms per sample, demonstrating its suitability for real-time applications. These results confirm that CBANet provides an effective balance between performance and computational efficiency.

\subsection{Evaluation at Session-Level and Driver-Level}

To assess the generalisation capability of the proposed model under realistic conditions, we evaluate CBANet at both the session level and the driver level, reducing bias from random data splitting.

At the \textbf{session level}, entire driving sessions are assigned exclusively to the training, validation, or testing sets, ensuring that no temporal segments from the same session appear across different splits. This setup prevents information leakage and evaluates robustness to temporal variations and intra-driver behavioural changes. As shown in Table~\ref{tab:generalisation_results}, CBANet maintains strong performance, demonstrating its ability to generalise across unseen driving sessions.

At the \textbf{driver level}, a stricter leave-one-driver-out protocol is adopted, where all sessions from unseen drivers are excluded during training. This setting evaluates generalisation to new drivers by eliminating driver-specific information leakage. CBANet maintains strong performance under this protocol, indicating that it learns driver-independent representations rather than overfitting to individual driving styles.

Overall, these results confirm that the proposed model generalises effectively across both unseen sessions and unseen drivers, supporting its suitability for real-world deployment.

\begin{table}[!t]
\centering
\caption{Session-Level and Driver-Level Evaluation Results}
\label{tab:generalisation_results}
\renewcommand{\arraystretch}{0.90}
\begin{tabular}{lccccc}
\toprule
\textbf{Setting} & \textbf{Accuracy} & \textbf{F2 (W)} & \textbf{F2 (M)} & \textbf{ROC-AUC} \\
\midrule
Session-Level & 0.9446 & 0.9446 & 0.8791 & 0.9929 \\
Driver-Level  & 0.9521 & 0.9519 & 0.8790 & 0.9934 \\
\bottomrule
\end{tabular}
\vspace{-1.5mm}
\end{table}

\subsection{Ablation Study}
\label{subsec:ablation}

To validate the effectiveness of the key design choices in the proposed framework, we conduct a series of ablation experiments focusing on both the network architecture and the loss function configuration, and the results are summarized in Tables~\ref{tab:ablation_architecture} and~\ref{tab:ablation_gamma}. First, to assess the contribution of different temporal modelling strategies, we evaluate multiple architectural variants, including standalone CNN, standalone LSTM, CNN+LSTM, LSTM with attention, and CNN with attention. These variants are compared under the same training and evaluation protocol to isolate the impact of architectural design on detection performance. In contrast, simply adding an attention mechanism does not lead to consistent performance gains and, in several cases, even results in a slight performance drop. This indicates that, for aggressive driving behaviour recognition, explicitly modelling the temporal dynamics of the signal is more important than relying on a generic attention mechanism alone.

 Second, to investigate the influence of the focal loss focusing parameter, we perform a sensitivity study by varying the value of $\gamma$ over a wide range, from $\gamma=0$ (equivalent to weighted cross-entropy) to larger values that increasingly emphasise hard samples. This set of experiments is designed to examine the trade-off between overall classification accuracy and recall-oriented performance on safety-critical events, particularly harsh braking.

The impact of the focal loss focusing parameter is analysed in Table ~\ref{tab:ablation_gamma}. It can be observed that setting $\gamma=0$, corresponding to a class-weighted cross-entropy loss, achieves the best overall trade-off across all evaluation metrics. As $\gamma$ increases from 0, performance consistently degrades, particularly in recall-oriented metrics. This behaviour can be attributed to the interaction between multiple imbalance mitigation strategies. Since the training pipeline already incorporates controlled SMOTE-based oversampling and class weighting, the effective class distribution becomes significantly more balanced. Under these conditions, additional focal modulation over-emphasises hard samples and introduces optimization instability, leading to reduced generalisation performance.

These results suggest that focal loss is most beneficial when used in isolation under severe imbalance, but may become redundant or even detrimental when combined with strong resampling and weighting strategies. In practice, we therefore adopt $\gamma=0$ (i.e., weighted cross-entropy) for all final experiments.


\begin{table}[!t]
\centering
\caption{Ablation Study on Network Architecture}
\label{tab:ablation_architecture}
\renewcommand{\arraystretch}{0.92} 
\begin{tabular}{lcccc}
\toprule
\textbf{Configuration} & \textbf{Accuracy} & \textbf{Macro F1} & \textbf{Macro F2} & \textbf{ROC-AUC} \\
\midrule
CNN              & 0.9159 & 0.8126 & 0.7927 & 0.9710 \\
LSTM             & 0.9198 & 0.8365 & 0.8426 & 0.9754 \\
CNN + LSTM       & 0.9251 & 0.8355 & 0.8097 & 0.9748 \\
LSTM + Attention & 0.8854 & 0.7923 & 0.8222 & 0.9751 \\
CNN + Attention  & 0.9057 & 0.7717 & 0.7346 & 0.9694 \\
\midrule
\textbf{CBANet} & \textbf{0.9585} & \textbf{0.9094} & \textbf{0.9067} & \textbf{0.9928} \\
\bottomrule
\end{tabular}
\vspace{-1.5mm}
\end{table}

\begin{table}[!b]
\centering
\caption{Ablation Study on Focal Loss Focusing Parameter $\gamma$}
\label{tab:ablation_gamma}
\renewcommand{\arraystretch}{0.92} 
\begin{tabular}{lcccc}
\toprule
\textbf{Configuration} & \textbf{Accuracy} & \textbf{Macro F1} & \textbf{Macro F2} & \textbf{ROC-AUC} \\
\midrule
$\gamma = 0.0$ & \textbf{0.9585} & \textbf{0.9094} & \textbf{0.9067} & \textbf{0.9928} \\
$\gamma = 1.0$ & 0.9298 & 0.8468 & 0.8277 & 0.9772 \\
$\gamma = 2.0$ & 0.9311 & 0.8478 & 0.8298 & 0.9765 \\
$\gamma = 3.0$ & 0.9120 & 0.7895 & 0.7540 & 0.9724 \\
$\gamma = 4.0$ & 0.9081 & 0.8088 & 0.7996 & 0.9619 \\
$\gamma = 5.0$ & 0.9119 & 0.8016 & 0.7814 & 0.9665 \\
\bottomrule
\end{tabular}
\vspace{-1.5mm}
\end{table}

\begin{figure}[!t]
\centering
\begin{tikzpicture}
\begin{axis}[
    width=0.95\columnwidth,        
    height=0.65\columnwidth,       
    ymin=0.75,
    ymax=0.93,
    ytick={0.75,0.79,0.83,0.87,0.91},
    ymajorgrids,
    ylabel={Macro F2},
    xlabel={Window Size $W$},
    xtick={1,2,3,4,5},
    legend columns=3,
    legend cell align={left},
    legend style={font=\small, at={(0.5,1.05)}, anchor=south},
    tick label style={font=\small},
    label style={font=\small},
]

\addplot[
    draw={rgb,1:red,0.0;green,0.45;blue,0.74},
    thick, mark=o
] coordinates {(1,0.8659) (2,0.8931) (3,0.9083) (4,0.8825) (5,0.9000)};

\addplot[
    draw={rgb,1:red,0.85;green,0.33;blue,0.10},
    thick, mark=square*
] coordinates {(1,0.8487) (2,0.8689) (3,0.8809) (4,0.8903) (5,0.8850)};

\addplot[
    draw={rgb,1:red,0.47;green,0.67;blue,0.19},
    thick, mark=triangle*
] coordinates {(1,0.8266) (2,0.8614) (3,0.8522) (4,0.8702) (5,0.8700)};

\addplot[
    draw={rgb,1:red,0.64;green,0.08;blue,0.18},
    thick, mark=diamond*
] coordinates {(1,0.7938) (2,0.8375) (3,0.8508) (4,0.8355) (5,0.8598)};

\addplot[
    draw={rgb,1:red,0.49;green,0.18;blue,0.56},
    thick, mark=+
] coordinates {(1,0.7625) (2,0.7913) (3,0.8157) (4,0.8359) (5,0.8457)};

\legend{$H=0$, $H=0.5$, $H=1.0$, $H=1.5$, $H=2.5$}

\end{axis}
\end{tikzpicture}
\caption{Macro F2 performance versus window size for different prediction horizons.}
\label{fig:Windows}
\end{figure}
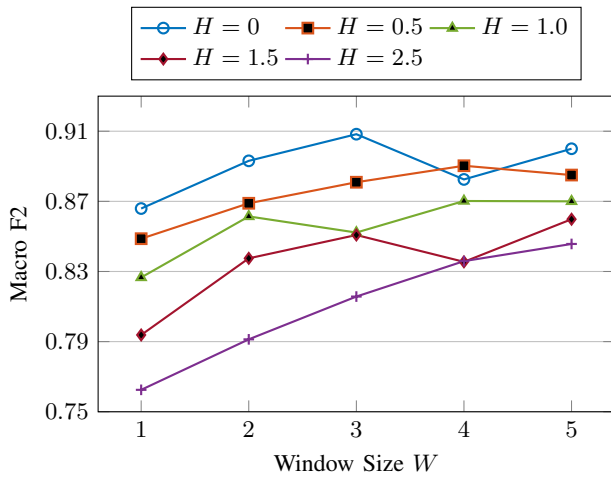

To analyse the sensitivity of the proposed model to temporal settings, we evaluate different input window sizes and prediction horizons on the real-world driving dataset. Specifically, the window length is set to $w \in \{1,2,3,4,5\}$ seconds, and the prediction horizon is varied over $H \in \{0.0, 0.5, 1.0, 1.5, 2.0\}$ seconds.
The detailed results of different parameters are presented in Fig~\ref{fig:Windows}. It can be observed that window sizes in the range of 3–5 seconds consistently yield stronger performance, while increasing the prediction horizon leads to a gradual but controlled performance degradation, indicating that the proposed model remains robust across varying temporal horizons.

\section{Conclusion}
\label{sec:Conclusion}
This paper proposed an enhanced, deep learning framework leveraging physically interpretable vehicle dynamics features for aggressive driving detection. By combining interpretable dynamic features, and imbalance-aware training, the approach overcomes key limitations of conventional data-driven methods. Experiments on a challenging naturalistic driving dataset show consistent improvements over strong baselines, especially in detecting rare but critical aggressive events. Specifically, CBANet achieves absolute gains of approximately 1.7\% in weighted F2 over the strongest deep sequential models and more than 6\% over graph-based approaches, while attaining the highest accuracy (0.9585) and ROC-AUC (0.9928). Future work will focus on multimodal extension and real-time edge deployment.

\bibliographystyle{IEEEtran}
\bibliography{references}

\end{document}